\pdfoutput=1

\documentclass[11pt]{article}

\usepackage{ACL2023}

\usepackage{times}
\usepackage{latexsym}
\usepackage{tabularx}
\usepackage{tablefootnote}
\usepackage{multirow}
\usepackage{graphicx}
\usepackage{amssymb}
\usepackage{float}

\usepackage[T1]{fontenc}

\usepackage[utf8]{inputenc}

\usepackage{microtype}

\usepackage{inconsolata}

\usepackage{enumitem}
\setitemize{noitemsep,topsep=0pt,parsep=0pt,partopsep=0pt}

\usepackage{listings}
\newcommand{\dagsuper}[0]{\textsuperscript{\textdagger}}
\newcommand{\starsuper}[0]{\textsuperscript{*}}

\title{FIREBALL: A Dataset of \textit{Dungeons and Dragons} Actual-Play with Structured Game State Information}

\author{Andrew Zhu$^1$, Karmanya Aggarwal$^1$, Alexander Feng$^1$,\\
 {\bf Lara J. Martin}$^2$\thanks{~~Work done while at the University of Pennsylvania.} , \and {\bf Chris Callison-Burch$^1$}\\
        $^1$University of Pennsylvania \\
        $^2$University of Maryland, Baltimore County \\
       {\tt \{andrz, karmanya, ahfeng, ccb\}@seas.upenn.edu},      {\tt laramar@umbc.edu}
}

\begin{document}

\maketitle
\begin{abstract}
Dungeons \& Dragons (D\&D) is a tabletop roleplaying game with complex natural language interactions between players and hidden state information. Recent work  has shown that large language models (LLMs) that have access to state information can generate higher quality game turns than LLMs that use dialog history alone.
However, previous work used game state information that was heuristically created and was not a true gold standard game state.
We present FIREBALL, a large dataset containing nearly 25,000 unique sessions from real D\&D gameplay on Discord with true game state info.  We recorded game play sessions of players who used the \textit{Avrae} bot, which was developed to aid people in playing D\&D online, capturing language, game commands and underlying game state information.  We demonstrate that FIREBALL can improve natural language generation (NLG) by using Avrae state information, improving both automated metrics and human judgments of quality. Additionally, we show that LLMs can generate executable Avrae commands, particularly after finetuning. 
\end{abstract}

\section{Introduction}

Dungeons \& Dragons (D\&D) \cite{Gygax1974} is a tabletop roleplaying game in which players assume the roles of characters in a fantasy adventure. Play is conducted primarily through natural language, with players roleplaying as their characters and describing their actions. Meanwhile another player, the Dungeon Master (DM), controls the fictional story world: setting obstacles, goals, and adventures, controlling monsters, and interpreting players' actions in the context of the rules of the game. Although the DM makes a lot of the final decisions, the game is ultimately a collaborative storytelling experience.

\begin{figure}[t]
\centering
\includegraphics[width=0.5\textwidth]{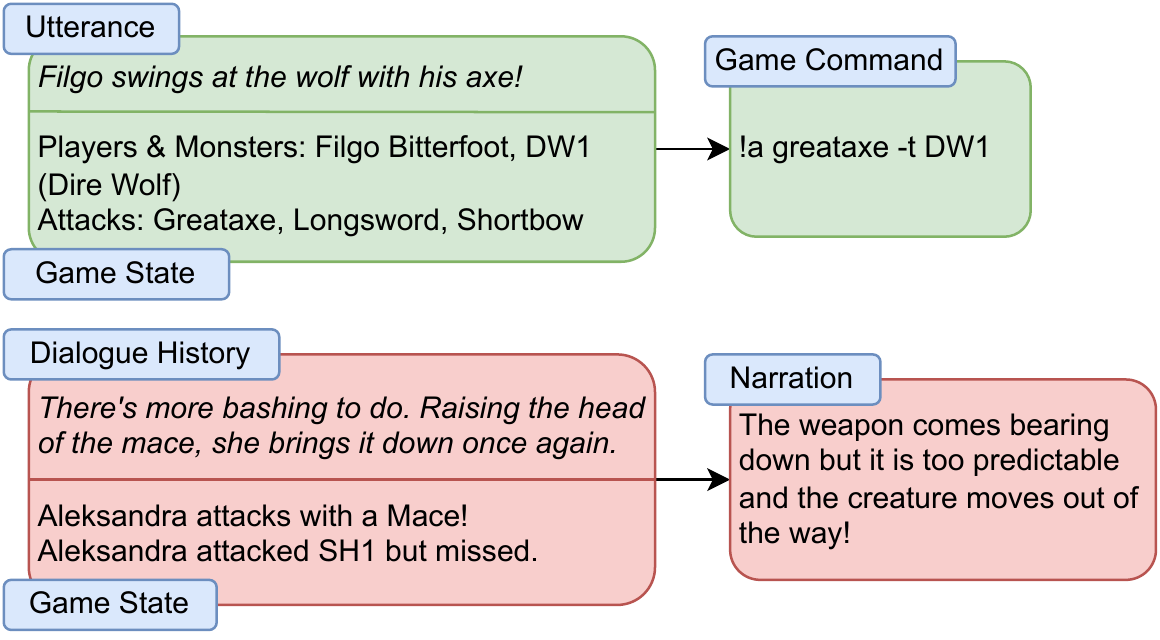}
\caption{Examples of our Utterance to Command task (top), which takes in an utterance and a game state to produce an Avrae command, and State to Narration task (bottom), which produces a narration given a dialogue history and game state information. }
\label{fig:sta_nar_ex}
\end{figure}

Due to its use of natural language as actions, each individual player must maintain a personal understanding of the game world which they build from conversational history and using Theory of Mind \cite{Martin2018, Zhou2022}.
The natural language action space also means that the AI needs the ability to adequately perform tasks such as language generation, language understanding, and planning \cite{callison-burch2022dungeons}.

Although AI's capabilities in this space are still nascent, \citet{callison-burch2022dungeons} have shown that D\&D dialog can be improved by adding state information into the input of a large language model (LLM). However, the state information presented in that work was heuristically created using regular expressions and machine learning classifiers.  Thus it cannot be considered true ground truth state information. Our work is unique because it provides true ground truth state information.

We use this data for two tasks: \textit{Utterance to Command} and \textit{State to Narration}.
In the first task, a model is given a game state and turn of the game (roleplay in natural language), and must predict the corresponding command that matches the \textit{intent} of the roleplay.  
The second task is a constrained creative natural language generation task: given a state change resulting from a command execution, generate a narration that describes the results.
 Figure \ref{fig:sta_nar_ex} demonstrates both tasks.


Our contributions are as follows:
\begin{itemize}
\item We present FIREBALL---a dataset of over 8M gameplay utterances, 2.1M commands, 1.2M gameplay states, and 160K unique actors (player \& non-player characters)\footnote{Our FIREBALL dataset is available here: \url{https://github.com/zhudotexe/FIREBALL}}. This is the first dataset of this size that includes detailed game state and character information for each scenario.
\item We show that large language models such as GPT-3, can extract relevant information from natural language in order to produce commands that are capable of being run by the game environment.
\item We demonstrate that LLMs, when finetuned on this dataset, can generate more grounded narrative text compared to language models tuned without game state information.
\end{itemize}




By incorporating structured state information into language understanding and NLG tasks, we hope to help pave the way toward more ambitious creative generation goals, such as consistent long-form narrative generation and video games that can convert language input into discrete game actions or generate narrations and dialogues based on the game state.

\section{Related Work}




Previous papers have outlined the challenges of Dungeons \& Dragons as an AI problem and examined various aspects of the game \cite{Ellis2017,Martin2018}.
Subsequently, a number of datasets in the D\&D space have been created \cite{louis-sutton-2018-deep,rameshkumar-bailey-2020-storytelling,si-etal-2021-telling,callison-burch2022dungeons,papazov2022using}, but these datasets either do not include game state information or include only an inexact game state, which lacks grounded, verified attributes, such as those included in our dataset.
Others have looked at using AI for subsets of D\&D gameplay, such as generating spell descriptions \cite{Newman2022} or simulating combat \cite{GamaVilaNova2019}.

In addition to the gold-labelled game state, we include all of the attributes from the following papers within our FIREBALL dataset.
\citet{louis-sutton-2018-deep} present a dataset pairing independent roleplay utterances with narrative descriptions of the characters that said them, and show that these descriptions influence the style of their utterances. More recently, \citet{callison-burch2022dungeons} frame D\&D as a dialogue challenge, approaching the challenge of generating cohesive and interesting gameplay utterances given the conversational history and a predicted game state. \citet{papazov2022using} present a pilot study on understanding player intent and translating it to a high level game action, represented as an Avrae command. Their workshop paper examines similar tasks, but studies only 100 hand-annotated events.  

Outside of the D\&D domain, the task of translating natural language into a bash command \cite{lin-etal-2018-nl2bash} or regex \cite{locascio-etal-2016-neural} is similar to our Utterance to Command task, although these languages do not require conditioning on a context. For another natural language-centric game,  Diplomacy, \citet{cicero-diplomacy} demonstrate the importance of providing game state information to language models. 

\begin{figure}[t]
\centering
\includegraphics[width=\linewidth]{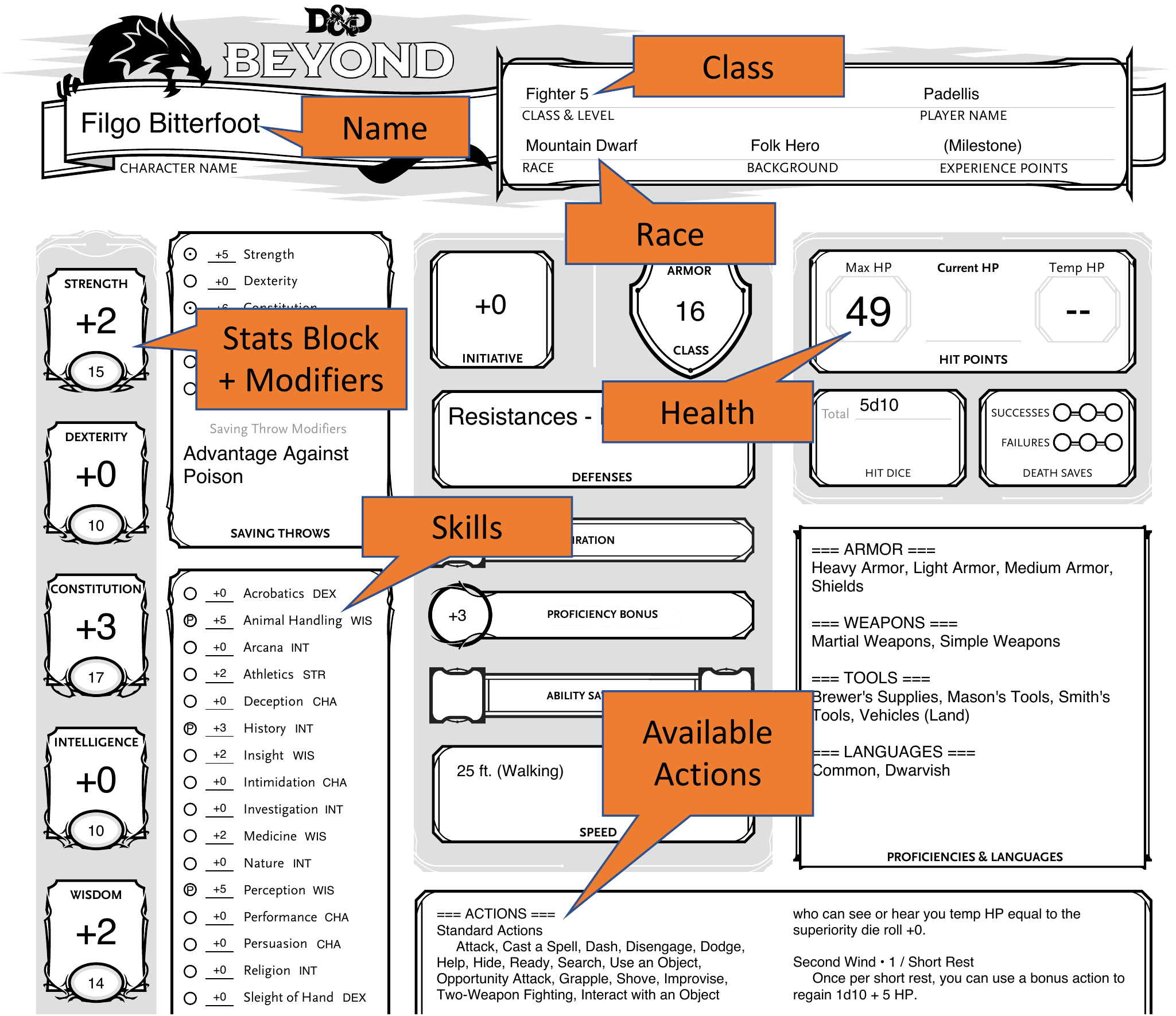}
\caption{A snippet of a Character Sheet for the character Filgo Bitterfoot. Players use pages like this to keep track of information about their D\&D characters.  We capture all this information in our state, in addition to information related to combat. Visit our repo (\url{https://github.com/zhudotexe/FIREBALL}) for the full document.}
\label{fig:character-sheet}
\end{figure}

\section{Playing D\&D Using Avrae}
In Dungeons \& Dragons, a group of players each create and play as a character.
Characters have classes (such as wizard or barbarian), fantasy races (such as Elves, Gnomes, and Dragonborn),  hit points (denoting their health), statistics that govern how well they can do certain actions, and inventories of items (armor, weapons, potions, etc).  These game state elements are stored on a character sheet (Figure \ref{fig:character-sheet}).  One player takes on the role of the Dungeon Master (DM). This special player creates the world in which the story is told, role plays all of the characters the other players interact with, and acts as arbiter of the rules of the game.

There are two main modes of gameplay --in-combat and out-of-combat-- which have different styles of play.
In-combat play simulates battles between characters and monsters, and involves turn taking and tracking stats. Out-of-combat play is characterized by freeform collaborative storytelling.
Both elements of the game involves rolling dice to determine the success of players' actions (like attempting to attack a monster). 
The die's outcome is then modified with a stat that represents the character's skill in performing that particular action (e.g., +3 for acrobatics).  If the dice roll plus the modifier was above a threshold that the rules or the DM determines (called a difficulty class, or DC), then the action succeeds.  Otherwise it fails. 
The DM then narrates the results. 

For example, if an attack hits a monster, the DM might narrate the player's blow and how the monster reacts, taking into account information like the attack's damage type and how many hits the monster has taken previously.


Traditionally, D\&D is played in person with characters' stats written out on physical character sheets and monster stats referenced from books containing hundreds of prewritten "stat blocks". To track a stat that changes frequently like hit points, players and DMs use paper and pencil or whiteboards, performing math in their head and writing over the previous value when it changes.
Some players also use maps and miniatures to track where characters and monsters are located relative to one another and to aid immersion in the game world. Since the beginning of the pandemic, a large number of groups have moved online using tools like Discord (a messaging program), virtual tabletops that simulate maps, and game state trackers like Avrae rather than physical mediums.

Avrae is a Discord bot that was designed to help people play D\&D online.  It allows players to import their character sheets, allows DMs to access a database of monsters, and simulates dice rolls. During combat, Avrae tracks the game state. This state contains detailed information including the list of participants in the battle, their stat blocks, their current HIT points, and their available actions.  Avrae allows players to execute commands representing their characters' actions.  It performs a simulated dice roll, adds the player's modifiers, and determines the success or failure of the roll.  Avrae then updates the game state, adjusting things like hit points, and turn tracking.

A simplified example of interacting with Avrae might look like the following example:
\begin{quote}
    \textbf{Player}: Filgo crouches down in the bush, loosing an arrow at the dire wolf charging towards him.

    \textbf{Player}: \verb|!attack longbow -t DW1|

    \textbf{Avrae}: (Rolls dice and displays the results of the attack and damage dealt, including the new health of the dire wolf.)

    \textbf{Dungeon Master}: Your arrow flies true and the beast lets out a shrill howl as it pierces its matted fur. It's low on health now, so on its turn it'll retreat.
\end{quote}
In actual play, an average of 3-8 players (including the DM) take turns interacting with Avrae.
By instrumenting Avrae to record these commands and messages before and after a user's inputted command, we collect a rich set of structured gameplay. Appendix \ref{sec:app-events} contains a full list of recorded events and their descriptions.

\section{Dataset}

We worked with the developer of Avrae to instrument it to collect game transcripts and log game state information.  The data collection was approved by Wizards of the Coast, the game company that owns D\&D and Avrae, as well as by our institution's IRB and the Bot Safety team at Discord. Any players who participated in our study provided their informed consent. 

\label{dataset}

\subsection{Data Collection}
Participants were recruited from English-speaking "play-by-post" D\&D Discord servers, where players and Dungeon Masters play by taking turns posting in a Discord text channel to describe their moves. To accomplish this, we made a website explaining the study, and recruited server admins to review the study and opt-in to participate by announcing our research study on the Avrae Discord server. When players began combat in an opted-in server, they were presented with a message informing them of the recording. Players could opt-out individual combat instances, server admins could stop all recording on a server at any time, and we provided a public form for requests to delete data.
For each actor (player or monster), we record a detailed state; Table \ref{table:char-attrs} in the Appendix lists all available attributes and their potential relevance to NLG tasks. Similarly, recorded actions include the detailed results of each dice roll, such as whether a given attack hit its target or a spell succeeded (a list of all action attributes is available in Appendix \ref{sec:action-events}). 

In the following sections, we refer to our data as triples consisting of a command and its corresponding state change, any relevant utterances before the command ("preceding" utterances), and any relevant utterances after the command ("following" utterances). Each of these commands corresponds to an action that an actor in combat can take, such as attacking with a weapon, casting a spell, or using a special ability.

\subsection{Utterance-Action Alignment}
\label{sec:alignment}
To align utterances with their corresponding state changes, we match each utterance with its chronologically nearest state change, and tag all utterances that occur chronologically before their corresponding state change as an utterance motivating the command (the "preceding" utterances), and all utterances that occur chronologically after their corresponding state change as narration of the state change (the "following" utterances). These alignments create a prototypical triple as described above. Within each triple, we discard any utterance containing less than five words.

\subsection{Authorship Filtering}

Within each triple, we identify the user who issued the commands, and the Dungeon Master hosting the combat. We discard any utterances within each triple which are not authored by one of these users. Additionally, we discard any triple where the commands originate from multiple different actors, which may occur if a single user is controlling multiple different creatures in a group. Finally, we discard any triple which has neither any "preceding" utterances nor "following" utterances.

\subsection{IC/OOC Classification}
\label{sec:OOC}
We further distill the set of "following" utterances by training GPT-3 Ada \cite{GPT3} to distinguish between "in-character" (IC) utterances from "out-of-character" (OOC) utterances. In-character utterances are what the player says speaking as their character or to describe their character's actions. They might look like this:

\begin{quote}
    Filgo puts a hand on his axe, uneasy after the shaking he'd felt from the ground.

    "Is someone there?"
\end{quote}

Meanwhile, out-of-character utterances occur across players/the DM when not speaking as any particular character. This dialog might be to discuss rules or strategy, or might be unrelated to gameplay entirely. Out-of-character utterances might look like this:

\begin{quote}
    How much health do you have left?
    
    I'll move back 30 feet after.

    BRB, going to the bathroom.
\end{quote}


To distinguish between these categories, we fine-tuned a classifier that was pretrained on Giant in the Playground data \cite{callison-burch2022dungeons}, forum roleplay data which includes labels for in-character and out-of-character posts, on a hand-labelled set of 750 utterances randomly sampled from our dataset. The classifier achieved an accuracy of 94\% on a validation set of 125 utterances. We then applied the classifier to each utterance in our dataset and discarded any out-of-character utterances from the "following" set since in-character text is usually more interesting and evocative.
Finally, we also removed sections of utterances contained in parentheses, which usually indicate OOC speech, from the "following" set.

\subsection{Dataset Size}

Our dataset contains 25k unique combat scenarios, including 8M utterances from 3.6k unique authors covering 1.3M unique combat states. Table \ref{table:commands} contains a breakdown of the distribution of commands in our dataset, organized by command category.

\begin{table}[t]
\small
\centering
\begin{tabular}{l c l} 
 {\bf Type} & {\bf Invocations} & {\bf Example}\\
 \hline\hline
 Combat & 713,568 & \verb|!init next| \\
 Actions & 608,527 & \verb|!cast fireball| \\
 Custom & 313,898 & \verb|!map| \\
 Character & 97,033 & \verb|!game hp| \\
 Checks\tablefootnote{Checks and saving throws contained as part of an action are not included in this category.} & 95,413 & \verb|!check arcana| \\
 Dice Rolls & 76,990 & \verb|!roll 1d20| \\
 Other & 204,174 & \verb|!help| \\
 \hline
 \textbf{Total} & 2,109,603
\end{tabular}
\normalsize
\caption{The number of 
command invocations in the dataset, organized by command category.}
\label{table:commands}
\end{table}

\section{Utterance to Command Task}


Our first task aims to predict the game command that a player or Dungeon Master intended to use, given the utterances since their last turn in combat. To successfully predict a command from an utterance, a model must be able to predict the user's intent, which actors the user intended to target, and ground both these predictions in the game state. For example, in the scenario illustrated in Figure \ref{fig:sta_nar_ex}, a dwarf named Filgo is fighting a Dire Wolf. On his turn, his player narrates that Filgo attacks with his axe, then runs the command to target the monster with his attack. Notice how, in this example, the player references the target dire wolf by its creature type ("the wolf"), rather than its name in the game state ("DW1").

To accomplish this task, we provide the state information included in our dataset---namely the list of actors participating in combat and any information about those actors, such as their monster type and current hit points---to the models. The full prompt for the example mentioned above is available in Appendix \ref{sec:full-utt-cmd-prompt}.

After our distillation passes, our dataset contains 120,000 aligned utterance-command pairs. We examine the accuracy of predicted commands on both a token level and by injecting predicted commands into the Avrae system. Finally, we also examine the performance of models without game state information included to demonstrate the importance of the game state.

\subsection{Models}
We use GPT-3 \cite{GPT3} Davinci models (as of Dec. 2022) as a base. Finetuned models are using standard Davinci, while few-shot models use Davinci-002. For the Utterance to Command generation task, we evaluate four main treatments, \begin{itemize}
\item \textbf{FT + S: }The base model is finetuned on a sample of 30K examples from FIREBALL with state information presented in the prompt.
\item \textbf{FT:} The base model is finetuned on a sample of 30K examples from FIREBALL without any state information presented in the prompt.
\item \textbf{FS + S:} The base model is presented 3 exemplars (few-shot) sampled from FIREBALL with relevant state information.
\item \textbf{FS:} The base model is presented 3 exemplars sampled from FIREBALL without any state information.
\end{itemize}

\subsection{Evaluation}
Each of the generation tasks is evaluated independently. Since the command generation task is more akin to a structured generation task, we evaluate only on objective correctness rather than subjective quality of generated text. We evaluate the generated commands over four quantitative metrics: pass rate, unit tests, RougeL \cite{rouge} and Average Sentence Gleu (SGleu) \cite{mutton-etal-2007-gleu}.

We first seek to evaluate whether the generated command is a valid Avrae command by simply passing the command to Avrae and checking for successful execution. Similar to \citet{codex} we calculate a \textit{pass rate} metric that determines the proportion of generations that constitute valid Avrae commands. To calculate this metric, we have each model generate commands for 1000 utterances randomly sampled from a held out test set, and simply count the proportion of generations that Avrae is able to successfully execute. 

Second, we evaluate what proportion of generated commands would result in the desired state update through a number of hand-written \textit{unit tests}. Each test accepts a predefined combat state, an utterance, and the corresponding model-generated command and validates assertions on the combat state update. We took 10 common scenarios seen in D\&D for these unit tests, generating 10 commands for each scenario-model pair. Since these generated commands sometimes have repeats, we take the $n$ unique commands from the set of generations and validate which proportion of the generated samples pass the handwritten unit tests by running assertions on the updated combat state after running the command through Avrae. 

Lastly, we perform a qualitative analysis to better understand the nature of the grounding. We took two popular spells (Bardic Inspiration and Fireball) and hand constructed a scenario for each. We then perturb the prompts that we present to the model to study the model's sensitivity to inputs.

\subsection{Results \& Discussion}
\begin{table}[]
\begin{tabular}{lcccc}
Model      & Pass Rate         & Unit Tests    & SGleu          & RougeL        \\ \hline \hline
FT+S       & \textbf{0.726} & \textbf{0.65} & \textbf{0.355} & 0.75 \\
FT    & 0.235          & 0.234         & 0.189          & 0.551         \\
FS+S   & 0.432          & 0.429         & 0.325          & \textbf{0.771}         \\
FS & 0.319          & 0.25          & 0.246          & 0.598        
\end{tabular}
\caption{Aggregated Results from the four models on the Utterance to Command Generation Task. SGleu refers to the average Sentence Gleu. }
\label{tab:utt-cmd-gen}
\end{table}
Table \ref{tab:utt-cmd-gen} displays the objective evaluations of the four treatments in the Utterance to Command generation task.  
We see that the model which was fine-tuned on FIREBALL with the state information (FT+S) significantly outperforms all other models (i.e., both models without the state information and the few-shot models). 
Within the perturbation results (detailed in Appendix \ref{perturbation-sec}), we notice that the FT+S model can accurately gauge the state of the actors in combat. For example, when asked to cast a Fireball at injured enemies, it successfully parses the prompt to find the subset of enemies that were injured and only targets them in the resulting command. Further, the model can accurately determine which of the prepared spells correspond to an utterance. For example, an utterance that would have generated the Fireball spell generates the spell Burning Hands instead if Fireball is removed from the prepared spell list.

\section{State to Narration Task}


In this task, we want to generate a narrative utterance describing the effects of a player's actions, given all of the state changes since the start of the player's turn in combat.

For example, in the scenario illustrated in Figure \ref{fig:sta_nar_ex}, a party is fighting a Sea Hag. On the cleric's turn, she attacks the hag with her mace, but misses. After seeing the result of her action, she narrates the miss, referencing the result as the hag dodging the attack. The full prompt for this example is available in Appendix \ref{sec:full-sta-nar-prompt}.

After our distillation and filter passes (Sections \ref{sec:alignment}-\ref{sec:OOC}), our dataset contains 43,000 aligned state-utterance pairs. To examine the importance of the game states provided in our dataset, we compare our results to methods that do not include the game state, such as dialog continuation \cite{callison-burch2022dungeons} and predicting the narration given only the command that was run \cite{papazov2022using}.

\subsection{Models}
We finetuned four GPT-3 \cite{GPT3} models on different data to determine the effect of state and dialog history inputs on generation.
Each model uses Davinci (as of Dec. 2022) as a base model, using 20,000 state-utterance pairs.

\begin{itemize}
\item \textbf{DIALOG: }Our first baseline model. This model is only given the last 5 messages of chat history, and fine-tuned to predict the next utterance that continues the dialog. It is not given any information about the game.
\item \textbf{COMMAND: }Our second baseline model. The model is only given the command that was run to take the player's action, and fine-tuned to predict the corresponding utterance.
\item \textbf{FIREBALL-SHORT:} Similar to DIALOG, but also contains the mechanical description of the action's results.
\item \textbf{FIREBALL-FULL:} All information given to FIREBALL-SHORT plus the full actor list, target list, and detailed attributes of the caster.
\end{itemize}

\subsection{Automated Evaluation}

\begin{table}[t]
\centering
\begin{tabularx}{\columnwidth}{X c c c} 
 Model & Perplexity & BERT & ROUGE \\

 \hline\hline
 DIALOG & 208.97 & 0.8458 & 0.1077 \\ \hline
 COMMAND & {\bf 156.98} & 0.8421 & 0.0919 \\\hline
 FIREBALL-SHORT & 202.39 & {\bf 0.8478} & 0.1087 \\\hline
 FIREBALL-FULL & 208.98 & 0.8476 & {\bf 0.1156} \\
 \hline\hline
 Human & 452.653 & N/A & N/A \\
\end{tabularx}
\caption{Perplexity, BERTScore, and ROUGE-1 scores of our models and human-written responses.}
\label{table:automated-sta-nar}
\end{table}

For the combat State to Narration generation task, we leverage standard text generation metrics: perplexity using a GPT-2 model \cite{GPT2} as a baseline, BertScore \cite{bertscore}, and ROUGE \cite{rouge}. All metrics aside from perplexity are calculated using the human narration as a reference. The results of our automated evaluation are available in Table \ref{table:automated-sta-nar}. 

We note that automated metrics are not particularly suited for evaluation of creative natural language generation. Perplexity is a measure of how "unexpected" a sequence is to a language model, which does not directly correlate with the \textit{quality} of creative generation. Furthermore, BertScore and ROUGE evaluate similarity to a reference, which is an imperfect fit for our task where two narrations can differ substantially yet both be of high quality.  These limitations are evident in the disparity in results between automated and human evaluation, which is expected given previous work that reached similar conclusions \cite{sagarkar-etal-2018-quality, delucia-etal-2021-decoding}.

\subsection{Human Evaluation}
We also perform a human evaluation to assess the quality of the generated utterances. In total, we recruited 45 evaluators from the Avrae user base based on their experience with Avrae and D\&D. All evaluators had played D\&D using Avrae before: 37 had used Avrae for over a year and 37 had been the Dungeon Master of a game using Avrae. Evaluators were rewarded with a set of codes for digital goods on D\&D Beyond with a market value of \$36 for completing the rating task.

We provided each evaluator a version of the context that was provided to the models: the last fifteen messages sent in a channel, the casting actor and their description, a list of actors in combat, and the current state of those actors. Along with each context, we provided one generated utterance from each model along with the true utterance sent by a human.
The evaluators were asked to rate each output along three dimensions, following the evaluation procedure used for the Meena LM \cite{MeenaLM} and D\&D Dialogue dataset \cite{callison-burch2022dungeons}: 
\begin{itemize}
    \item Does the response make sense? (yes/no) 
    \item Is the response specific? (yes/no) 
    \item How interesting is the response? (10 point scale)
\end{itemize}

Each evaluator rated 3 to 7 scenarios randomly drawn from a set of 75, with at least 3-way redundancy for each scenario. The full annotator instructions and a mockup of the annotation interface are given in Appendix \ref{sec:eval-interface}.

\subsection{Results \& Discussion}

\begin{table}[t]
\small
\centering
\begin{tabularx}{\columnwidth}{X c c c} 
 Model & Sense & Specific & Interest \\
 \hline\hline
 DIALOG & 0.36 & 0.27 & 4.27 \\ 
 COMMAND & 0.41 & 0.37 & 4.72 \\
 FIREBALL-SHORT & 0.52 & {\bf 0.48} & {\bf 4.98} \\
 FIREBALL-FULL & {\bf 0.55} & 0.47 & 4.6 \\
 \hline
 Human & 0.54 & 0.48 & 4.91 \\
\end{tabularx}
\normalsize
\caption{Average sense, specific, and interestingness scores of our models and human-written responses.}
\label{table:human-sta-nar}
\end{table}

\begin{table*}[h]
\small
\centering
{\setlength{\extrarowheight}{.5em}
\begin{tabularx}{\textwidth}{l X} 
 Last Human Utterance & "Dogs... so many of them making it more annoying then anything." Umbrage didn't want to waste anything big. So he pulled out his pipes, taking the attack of opportunity as he would start to play. "Fear me!" \\
 Command Run & \verb|!i cast "Cause Fear" -t dd1 -t dd3 -t dd4| [...] \\
 \hline
 {\bf Model} & {\bf Narration} \\
 \hline\hline
 DIALOG & The music was certainly not pleasant, but it seemed to have an effect on the dogs. They became panicked for a moment, but were able to maintain their aggression. It was just enough for the party to act. \\
 COMMAND & Cuco proceeds to try and scare the ducks. So far it wasn't working. \\
 FIREBALL-SHORT & As the first dog fell, the others seemed to flee once more in pure terror of the seemingly unending tunes of the pipes. \\
 FIREBALL-FULL & The dogs were in the middle of a panic. But still they fought on! \\
 Human (Original Narration) & Moving back into the melee with the one he was originally fighting. As he drew ready.
\end{tabularx}
}
\normalsize
\caption{Examples of predicted completions from each model given the last human utterance, the command that was run, or both. In this scenario, a Dragonborn Bard named Umbrage has just cast the Cause Fear spell against a pack of Death Dogs. As a result of the command being run, three of the dogs fail to save and are afraid; four resist the effect.}
\label{table:examples-sta-nar}
\end{table*}

The results of our human evaluation are tabulated in Table \ref{table:human-sta-nar}. Both FIREBALL models outperform the baseline models in sensibility and specificity by an average of 15 percentage points (significant to $p < 0.01$), and on average perform similarly to a human ($p > 0.5$). A detailed analysis of significance can be found in Appendix \ref{sec:human-eval-sig}. Generally, models that were aware of the game context (including COMMAND) were more interesting than the model tasked with simply continuing the chat history, and comparable to human performance.

It may seem unusual for the human performance to be so low, only making sense to an experienced D\&D player about 50\% of the time.
One explanation could be that human-written responses were more likely to refer to background knowledge not provided in the model's context, and therefore may have caused raters to mark the response as nonsensical.
We compiled qualitative feedback from our evaluators to provide some insight into why this may be, as well as identify some common failure cases of our models. We summarize some of the reoccurring themes here.

\paragraph{Removing Player Agency.} The most common theme among evaluators' feedback was that they did not want the model to take control of their character away from the player. One evaluator noted that "\emph{several [narrations] had player characters acting or speaking in response to the action. That's something I would never want a human DM doing unprompted and it might be frustrating to have the bot look like it's trying to control what my character does or says.}"
This problem extended to the Dungeon Master's role, as well; one evaluator mentioned that some AI responses would specify creatures' movement and "drag [their] encounters down the hallway."

Player agency is an especially challenging aspect of the game to maintain while training language models with real player utterances, as the training data available naturally makes decisions for and speaks as the player. 
Multiple evaluators suggested that language models should be used to provide assistance and inspiration for players rather than directly acting on behalf of the player, allowing a human to make the final decision for their characters.

\paragraph{Hit Point Reasoning.} Another common failure case was models' tendency to describe any damage to a target, regardless of the target's true remaining health, as a kill. According to one evaluator, "\emph{I ran into a scenario where the text was absolutely great, but incredibly wrong. A player scored a critical hit on a Mind Flayer with a bow, and left the mind flayer bloodied. The long text stated the scenario beautifully, even got the weapon right... but said it killed the Mind Flayer.}"

We noticed this failure case was more prevalent among the DIALOG and COMMAND models, demonstrating that our state-aware models were able to appropriately ground the generated narrations.
Nevertheless, even the state-aware models often made this mistake. 
We have seen in our dataset that humans tend to only narrate critical portions of a fight, like when an enemy is killed.
This reporting bias creates a disproportionately large number of narrations in our training set that reference a creature being killed. 
Therefore, the models trained on this data skew towards narrating that monsters are killed.

Table \ref{table:examples-sta-nar} shows a scenario that exemplifies many of these aspects.
Without the guidance of previous dialog turns, the COMMAND model hallucinates names for the caster and the targeted creatures. The DIALOG, FIREBALL-SHORT, and FIREBALL-FULL models produce plausible next utterances based on the details provided in the player's utterance but exhibit some of the discussed failure cases: it acts as the Dungeon Master to narrate that the dogs have fled and it references the first dog dying, which it did not (in the game state, the dog still has a full 39/39 hit points).
For reference, the true next utterance as written by the player does not mention any effect of the spell, instead focusing on the character's movement. The full game state and chat history associated with this example is included in Appendix \ref{sec:sta-nar-example}.

\section{Conclusions}
We have demonstrated how the FIREBALL dataset can be used to predict game commands that correctly match a player's intent and generate cohesive and grounded narration from the results of a game action.
Our Utterance to Command model is capable of translating roleplay into game-specific actions and can aid novice users, reducing the amount of time players spend looking up documentation and allowing them to play the game more. 
Our State to Narration model can help inspire the Dungeon Master and take some of the cognitive load off of repetitive writing tasks, allowing them to focus on creating an enjoyable experience for the players.
FIREBALL opens the door to multiple exciting avenues of research, and we're excited to see how future work utilizes our unique dataset of state-augmented gameplay.

\section{Limitations}

Dungeons \& Dragons is a very complex game to capture completely, and there are certain aspects that FIREBALL does not take into account. For example, FIREBALL's scenarios are recorded independently of the overarching narrative context they take place in, do not record players' inventory, and do not account for any movement or placement on a map.
Our models are not able to play D\&D autonomously - but doing so is not the goal. Instead, D\&D models can be used to assist and inspire the humans playing.

Our models do not take into account the generation of profanity or sensitive topics; these were filtered out post-hoc. D\&D is a game played by players of all ages that often contains violent or profane descriptions, and unfiltered generations may be unsuitable for young players.  There are previous instances of roleplaying games that incorporate language models being used to generate sexual content\footnote{\url{https://www.theregister.com/2021/10/08/ai_game_abuse/}} that would require age restrictions and content warnings. 

GPT-3 may be prohibitively expensive for everyday use; in our experiments, we were unable to use the full set of data we had available for fine-tuning due to budget constraints.


\section*{Acknowledgements}

We would like to thank Pat Backmann and others at D\&D Beyond for supporting us with gift codes for human evaluators, access to the Avrae system, and generally being an enthusiastic proponent of our work.

Thank you to the members of the Avrae Development, Avrae AI Human Evaluation, and Northern Lights Province Discord servers for helping us evaluate our systems, your excitement about AI in D\&D, and initiative in offering us feedback! 

This research is based upon work supported in part by the DARPA KAIROS Program (contract FA8750-19-2-1004), the DARPA LwLL Program (contract FA8750-19-2-0201), the IARPA HIATUS Program (contract 2022-22072200005), and the NSF (Award 1928631 and Grant \#2030859 to the Computing Research Association for the CIFellows Project). Approved for Public Release, Distribution Unlimited. The views and conclusions contained herein are those of the authors and should not be interpreted as necessarily representing the official policies, either expressed or implied, of DARPA, IARPA, NSF, or the U.S. Government.

\bibliography{anthology,custom}

\begin{thebibliography}{23}
\expandafter\ifx\csname natexlab\endcsname\relax\def\natexlab#1{#1}\fi

\bibitem[{Bakhtin et~al.(2022)Bakhtin, Brown, Dinan, Farina, Flaherty, Fried,
  Goff, Gray, Hu, Jacob, Komeili, Konath, Kwon, Lerer, Lewis, Miller, Mitts,
  Renduchintala, Roller, Rowe, Shi, Spisak, Wei, Wu, Zhang, and
  Zijlstra}]{cicero-diplomacy}
Anton Bakhtin, Noam Brown, Emily Dinan, Gabriele Farina, Colin Flaherty, Daniel
  Fried, Andrew Goff, Jonathan Gray, Hengyuan Hu, Athul~Paul Jacob, Mojtaba
  Komeili, Karthik Konath, Minae Kwon, Adam Lerer, Mike Lewis, Alexander~H.
  Miller, Sasha Mitts, Adithya Renduchintala, Stephen Roller, Dirk Rowe, Weiyan
  Shi, Joe Spisak, Alexander Wei, David Wu, Hugh Zhang, and Markus Zijlstra.
  2022.
\newblock \href {https://doi.org/10.1126/science.ade9097} {Human-level play in
  the game of diplomacy by combining language models with strategic reasoning}.
\newblock \emph{Science}, 378(6624):1067--1074.

\bibitem[{Brown et~al.(2020)Brown, Mann, Ryder, Subbiah, Kaplan, Dhariwal,
  Neelakantan, Shyam, Sastry, Askell, Agarwal, Herbert-Voss, Krueger, Henighan,
  Child, Ramesh, Ziegler, Wu, Winter, Hesse, Chen, Sigler, Litwin, Gray, Chess,
  Clark, Berner, McCandlish, Radford, Sutskever, and Amodei}]{GPT3}
Tom Brown, Benjamin Mann, Nick Ryder, Melanie Subbiah, Jared~D Kaplan, Prafulla
  Dhariwal, Arvind Neelakantan, Pranav Shyam, Girish Sastry, Amanda Askell,
  Sandhini Agarwal, Ariel Herbert-Voss, Gretchen Krueger, Tom Henighan, Rewon
  Child, Aditya Ramesh, Daniel Ziegler, Jeffrey Wu, Clemens Winter, Chris
  Hesse, Mark Chen, Eric Sigler, Mateusz Litwin, Scott Gray, Benjamin Chess,
  Jack Clark, Christopher Berner, Sam McCandlish, Alec Radford, Ilya Sutskever,
  and Dario Amodei. 2020.
\newblock \href
  {https://proceedings.neurips.cc/paper/2020/file/1457c0d6bfcb4967418bfb8ac142f64a-Paper.pdf}
  {Language models are few-shot learners}.
\newblock In \emph{Advances in Neural Information Processing Systems},
  volume~33, pages 1877--1901. Curran Associates, Inc.

\bibitem[{Callison-Burch et~al.(2022)Callison-Burch, Singh~Tomar, Martin,
  Ippolito, Bailis, and Reitter}]{callison-burch2022dungeons}
Chris Callison-Burch, Gaurav Singh~Tomar, Lara~J. Martin, Daphne Ippolito, Suma
  Bailis, and David Reitter. 2022.
\newblock {Dungeons and Dragons as a Dialogue Challenge for Artificial
  Intelligence}.
\newblock In \emph{Conference on Empirical Methods in Natural Language
  Processing (EMNLP)}.

\bibitem[{Chen et~al.(2021)Chen, Tworek, Jun, Yuan, Pinto, Kaplan, Edwards,
  Burda, Joseph, Brockman, Ray, Puri, Krueger, Petrov, Khlaaf, Sastry, Mishkin,
  Chan, Gray, Ryder, Pavlov, Power, Kaiser, Bavarian, Winter, Tillet, Such,
  Cummings, Plappert, Chantzis, Barnes, Herbert-Voss, Guss, Nichol, Paino,
  Tezak, Tang, Babuschkin, Balaji, Jain, Saunders, Hesse, Carr, Leike, Achiam,
  Misra, Morikawa, Radford, Knight, Brundage, Murati, Mayer, Welinder, McGrew,
  Amodei, McCandlish, Sutskever, and Zaremba}]{codex}
Mark Chen, Jerry Tworek, Heewoo Jun, Qiming Yuan, Henrique Ponde de~Oliveira
  Pinto, Jared Kaplan, Harri Edwards, Yuri Burda, Nicholas Joseph, Greg
  Brockman, Alex Ray, Raul Puri, Gretchen Krueger, Michael Petrov, Heidy
  Khlaaf, Girish Sastry, Pamela Mishkin, Brooke Chan, Scott Gray, Nick Ryder,
  Mikhail Pavlov, Alethea Power, Lukasz Kaiser, Mohammad Bavarian, Clemens
  Winter, Philippe Tillet, Felipe~Petroski Such, Dave Cummings, Matthias
  Plappert, Fotios Chantzis, Elizabeth Barnes, Ariel Herbert-Voss,
  William~Hebgen Guss, Alex Nichol, Alex Paino, Nikolas Tezak, Jie Tang, Igor
  Babuschkin, Suchir Balaji, Shantanu Jain, William Saunders, Christopher
  Hesse, Andrew~N. Carr, Jan Leike, Josh Achiam, Vedant Misra, Evan Morikawa,
  Alec Radford, Matthew Knight, Miles Brundage, Mira Murati, Katie Mayer, Peter
  Welinder, Bob McGrew, Dario Amodei, Sam McCandlish, Ilya Sutskever, and
  Wojciech Zaremba. 2021.
\newblock \href {https://doi.org/10.48550/ARXIV.2107.03374} {Evaluating large
  language models trained on code}.

\bibitem[{DeLucia et~al.(2021)DeLucia, Mueller, Li, and
  Sedoc}]{delucia-etal-2021-decoding}
Alexandra DeLucia, Aaron Mueller, Xiang~Lisa Li, and Jo{\~a}o Sedoc. 2021.
\newblock \href {https://doi.org/10.18653/v1/2021.gem-1.16} {Decoding methods
  for neural narrative generation}.
\newblock In \emph{Proceedings of the 1st Workshop on Natural Language
  Generation, Evaluation, and Metrics (GEM 2021)}, pages 166--185, Online.
  Association for Computational Linguistics.

\bibitem[{Ellis and Hendler(2017)}]{Ellis2017}
Simon Ellis and James Hendler. 2017.
\newblock \href {https://doi.org/10.1109/MIS.2017.3121545} {{Computers Play
  Chess, Computers Play Go...Humans Play Dungeons \& Dragons}}.
\newblock \emph{IEEE Intelligent Systems}, 32(4):31–34.

\bibitem[{Gama Vila~Nova et~al.(2019)Gama Vila~Nova, Carneiro~Vital, and
  Vilhena Vieira~Lopes}]{GamaVilaNova2019}
João~Gabriel Gama Vila~Nova, Marcos~Vinícius Carneiro~Vital, and Roberta
  Vilhena Vieira~Lopes. 2019.
\newblock \href {https://doi.org/10.1007/978-3-030-30241-2_7} {{A Reliable
  Information Acquisition Model for D\&D Players}}.
\newblock In \emph{Progress in Artificial Intelligence}, {Lecture Notes in
  Computer Science}, page 73–85, Cham. Springer International Publishing.

\bibitem[{Gygax and Arneson(1974)}]{Gygax1974}
Gary Gygax and Dave Arneson. 1974.
\newblock \emph{{Dungeons \& Dragons}}.

\bibitem[{Kulshreshtha et~al.(2020)Kulshreshtha, Adiwardana, So, Nemade, Hall,
  Fiedel, Le, Thoppilan, Luong, Lu, and Yang}]{MeenaLM}
Apoorv Kulshreshtha, Daniel De~Freitas Adiwardana, David~Richard So, Gaurav
  Nemade, Jamie Hall, Noah Fiedel, Quoc~V. Le, Romal Thoppilan, Thang Luong,
  Yifeng Lu, and Zi~Yang. 2020.
\newblock Towards a human-like open-domain chatbot.
\newblock In \emph{arXiv}.

\bibitem[{Lin(2004)}]{rouge}
Chin-Yew Lin. 2004.
\newblock \href {https://aclanthology.org/W04-1013} {{ROUGE}: A package for
  automatic evaluation of summaries}.
\newblock In \emph{Text Summarization Branches Out}, pages 74--81, Barcelona,
  Spain. Association for Computational Linguistics.

\bibitem[{Lin et~al.(2018)Lin, Wang, Zettlemoyer, and
  Ernst}]{lin-etal-2018-nl2bash}
Xi~Victoria Lin, Chenglong Wang, Luke Zettlemoyer, and Michael~D. Ernst. 2018.
\newblock \href {https://aclanthology.org/L18-1491} {{NL}2{B}ash: A corpus and
  semantic parser for natural language interface to the linux operating
  system}.
\newblock In \emph{Proceedings of the Eleventh International Conference on
  Language Resources and Evaluation ({LREC} 2018)}, Miyazaki, Japan. European
  Language Resources Association (ELRA).

\bibitem[{Locascio et~al.(2016)Locascio, Narasimhan, DeLeon, Kushman, and
  Barzilay}]{locascio-etal-2016-neural}
Nicholas Locascio, Karthik Narasimhan, Eduardo DeLeon, Nate Kushman, and Regina
  Barzilay. 2016.
\newblock \href {https://doi.org/10.18653/v1/D16-1197} {Neural generation of
  regular expressions from natural language with minimal domain knowledge}.
\newblock In \emph{Proceedings of the 2016 Conference on Empirical Methods in
  Natural Language Processing}, pages 1918--1923, Austin, Texas. Association
  for Computational Linguistics.

\bibitem[{Louis and Sutton(2018)}]{louis-sutton-2018-deep}
Annie Louis and Charles Sutton. 2018.
\newblock \href {https://doi.org/10.18653/v1/N18-2111} {Deep dungeons and
  dragons: Learning character-action interactions from role-playing game
  transcripts}.
\newblock In \emph{Proceedings of the 2018 Conference of the North {A}merican
  Chapter of the Association for Computational Linguistics: Human Language
  Technologies, Volume 2 (Short Papers)}, pages 708--713, New Orleans,
  Louisiana. Association for Computational Linguistics.

\bibitem[{Martin et~al.(2018)Martin, Sood, and Riedl}]{Martin2018}
Lara~J. Martin, Srijan Sood, and Mark~O. Riedl. 2018.
\newblock \href {http://ceur-ws.org/Vol-2321/paper4.pdf} {{Dungeons and DQNs:
  Toward Reinforcement Learning Agents that Play Tabletop Roleplaying Games}}.
\newblock In \emph{Joint Workshop on Intelligent Narrative Technologies and
  Workshop on Intelligent Cinematography and Editing (INT-WICED)}, Edmonton,
  AB, Canada. http://ceur-ws.org.

\bibitem[{Mutton et~al.(2007)Mutton, Dras, Wan, and
  Dale}]{mutton-etal-2007-gleu}
Andrew Mutton, Mark Dras, Stephen Wan, and Robert Dale. 2007.
\newblock \href {https://aclanthology.org/P07-1044} {{GLEU}: Automatic
  evaluation of sentence-level fluency}.
\newblock In \emph{Proceedings of the 45th Annual Meeting of the Association of
  Computational Linguistics}, pages 344--351, Prague, Czech Republic.
  Association for Computational Linguistics.

\bibitem[{Newman and Liu(2022)}]{Newman2022}
Pax Newman and Yudong Liu. 2022.
\newblock \href {https://aclanthology.org/2022.games-1.7/} {{Generating
  Descriptive and Rules-Adhering Spells for Dungeons \& Dragons Fifth
  Edition}}.
\newblock In \emph{Games and Natural Language Processing Workshop at LREC},
  page 54–60, Marseille, France. European Language Resources Association.

\bibitem[{Papazov et~al.(2022)Papazov, Gill, Ferreiro, Zhu, Martin, and
  Callison-Burch}]{papazov2022using}
Stefan Papazov, Wesley Gill, Marta~Garcia Ferreiro, Andrew Zhu, Lara~J. Martin,
  and Chris Callison-Burch. 2022.
\newblock \href {https://openreview.net/forum?id=jQSStHwtmDN} {Using language
  models to convert between natural language and game commands}.
\newblock In \emph{The Third Wordplay: When Language Meets Games Workshop}.

\bibitem[{Radford et~al.(2019)Radford, Wu, Child, Luan, Amodei, and
  Sutskever}]{GPT2}
Alec Radford, Jeff Wu, Rewon Child, David Luan, Dario Amodei, and Ilya
  Sutskever. 2019.
\newblock Language models are unsupervised multitask learners.

\bibitem[{Rameshkumar and Bailey(2020)}]{rameshkumar-bailey-2020-storytelling}
Revanth Rameshkumar and Peter Bailey. 2020.
\newblock \href {https://doi.org/10.18653/v1/2020.acl-main.459} {Storytelling
  with dialogue: {A} {Critical Role Dungeons and Dragons Dataset}}.
\newblock In \emph{Proceedings of the 58th Annual Meeting of the Association
  for Computational Linguistics}, pages 5121--5134, Online. Association for
  Computational Linguistics.

\bibitem[{Sagarkar et~al.(2018)Sagarkar, Wieting, Tu, and
  Gimpel}]{sagarkar-etal-2018-quality}
Manasvi Sagarkar, John Wieting, Lifu Tu, and Kevin Gimpel. 2018.
\newblock \href {https://doi.org/10.18653/v1/S18-2024} {Quality signals in
  generated stories}.
\newblock In \emph{Proceedings of the Seventh Joint Conference on Lexical and
  Computational Semantics}, pages 192--202, New Orleans, Louisiana. Association
  for Computational Linguistics.

\bibitem[{Si et~al.(2021)Si, Ammanabrolu, and Riedl}]{si-etal-2021-telling}
Wai~Man Si, Prithviraj Ammanabrolu, and Mark Riedl. 2021.
\newblock \href {https://aclanthology.org/2021.sigdial-1.30} {Telling stories
  through multi-user dialogue by modeling character relations}.
\newblock In \emph{Proceedings of the 22nd Annual Meeting of the Special
  Interest Group on Discourse and Dialogue}, pages 269--275, Singapore and
  Online. Association for Computational Linguistics.

\bibitem[{Zhang et~al.(2019)Zhang, Kishore, Wu, Weinberger, and
  Artzi}]{bertscore}
Tianyi Zhang, Varsha Kishore, Felix Wu, Kilian~Q. Weinberger, and Yoav Artzi.
  2019.
\newblock \href {https://doi.org/10.48550/ARXIV.1904.09675} {Bertscore:
  Evaluating text generation with bert}.

\bibitem[{Zhou et~al.(2022)Zhou, Zhu, Hu, Pujara, Ren, Callison-Burch, Choi,
  and Ammanabrolu}]{Zhou2022}
Pei Zhou, Andrew Zhu, Jennifer Hu, Jay Pujara, Xiang Ren, Chris Callison-Burch,
  Yejin Choi, and Prithviraj Ammanabrolu. 2022.
\newblock \href {https://doi.org/10.48550/ARXIV.2212.10060} {An ai dungeon
  master's guide: Learning to converse and guide with intents and
  theory-of-mind in dungeons and dragons}.

\end{thebibliography}
\bibliographystyle{acl_natbib}

\newpage
\onecolumn
\appendix
\section{Example of Collected Character Information}
Here, Table \ref{table:char-attrs}, we provide a table of all of the state and character information that we collected from actual play. The table gives an example for the character Filgo Bitterfoot, a level 5 Dwarf Fighter. We also explain the function of each attribute within the rules of D\&D, as well as a possible use case for generative applications.

\begin{table*}[th!]
\small
\centering
\begin{tabularx}{\textwidth}{|l|X|X|X|} 
 \hline
 {\bf Attribute} & {\bf Example} & {\bf Description} & {\bf Use Case} \\
 \hline\hline
 Name & Filgo Bitterfoot & Character's name. & {Both commands and narration usually reference actors by their name.} \\
 \hline
 Class Level & Fighter 5 & Classes inform what types of actions the character is good at, while level is the amount of "experience" the character has. &{An actor's classes offer a general overview of their abilities.} \\
 \hline
 Stats & STR: 15; DEX: 10; CON: 17; INT: 10; WIS: 14; CHA: 10; Proficiency: +3 & \multirow{2}{\hsize}{Strength, dexterity, constitution, intelligence, wisdom, \& charisma. Being proficient in a stat or skill means that a bonus is added to their die outcome.} & \multirow{3}{\hsize}{Can be used to condition text generation; for example, a strong character's actions would likely be described differently from a  dexterous character.} \\
 \cline{1-2}
 Skills & Acrobatics +0; Animal Handling +5 (proficient);... && \\
 \cline{1-3}
 Saves & STR +5; DEX +0; CON +6; INT +0; WIS +2; CHA +0 & Saves are bonuses for when a character is defending against an attack. & \\
 \hline
 Resistances & Resistant to fire, poison & Cuts the amount of damage done or removes it entirely if they are \textit{immune}.&{Useful for game reasoning: don't use fire abilities against an immune target.} \\
 \hline
 Attacks & Frost Brand Scimitar (+5 to hit, 2d6+2 damage); Unarmed Strike (+5 to hit, 1d4+2 damage); ... & Weapons, bonuses for the ability to hit the target, and the amount of damage the target takes when hit. 2d6 indicates 2 6-sided dice. & \multirow{3}{\hsize}{These attributes provide a more detailed list of the actor's capabilities. For command prediction, the actor may only use an ability they actually have. For text generation, an ability list can help describe the actor's style (e.g. "an ice mage").} \\
 \cline{1-3}
 Spellbook & +5 Spell Bonus; DC 13; Spells: Fireball, Prestidigitation, ... & Bonus to see if a spell hits the target, the amount the target needs to beat when being hit by the spell in order to dodge it, a list of spells that the character can cast (each with different outcomes) & \\
 \cline{1-3}
 Actions\dagsuper & Rally; Second Wind; Interception; Action Surge; Disarming Attack; ... & Class-specific abilities.&\\
 \hline
 Custom Counters\dagsuper & Superiority Dice: 4/4; Second Wind: 1/1; Action Surge: 1/1 & Limitation on the number of times the character can use an action before resting. & {Certain abilities can only be used a certain number of times.} \\
 \hline
 Armor Class & 18 & How difficult the character is to hit. & \multirow{5}{\hsize}{Provides information about how dire a situation might be for more interesting text generation. Actors are also commonly referred to by a combination of these (e.g. "the prone dwarf").} \\
 \cline{1-3}
 Hit Points & 54 / 54 (0 temp) & How "healthy" a character is. Filgo has 54 points out of 54.&\\
 \cline{1-3}
 Effects\starsuper & Prone, Stunned & Combat-specific states that affect how attacks resolve. &\\
 \cline{1-3}
 Creature Type & Humanoid & Category of creature. Other examples are \textit{undead} and \textit{dragon}. & \\
 \cline{1-3}
 Race\dagsuper & Mountain Dwarf & Fantasy race that might provide extra bonuses or actions. & \\
 \hline
 Description\dagsuper & "Filgo is a level 5 Mountain Dwarf Fighter. He is 100 years old, 4' tall..." & Natural-language description of the character, how they look \& what they're like.& {Provides qualitative information about the actor useful for text conditioning, e.g. \citet{louis-sutton-2018-deep}.} \\
 \hline
\end{tabularx}

\normalsize
\caption{A list of common attributes associated with actors in FIREBALL. \dagsuper Only available for player characters. \starsuper Only available for actors in combat. Character sheet for this example actor available at \url{https://github.com/zhudotexe/FIREBALL/blob/main/Filgo_Bitterfoot.pdf}.}
\label{table:char-attrs}
\end{table*}

\section{Dataset Events}
\label{sec:app-events}

\begin{table}[h]
\centering
\small
\begin{tabularx}{\textwidth}{|l|c|X|} 
 \hline
 {\bf Event} & {\bf Count} & {\bf Description}\\
 \hline\hline
 \verb|message| & 8,012,706 & A user has sent a message in the recorded Discord channel. \\
 \hline
 \verb|command| & 2,109,603 & An Avrae command has successfully executed. Includes information about the game actor that ran the command. \\
 \hline
 \verb|combat_state_update| & 1,297,254 & Some element of the combat state has changed. Includes the new combat state. \\
 \hline
 \verb|automation_run| & 588,712 & An action has successfully executed. Includes information about all the rolls in and result of the action, including successes and failures of attacks, saves, etc. \\
 \hline
 \verb|alias_resolution| & 325,860 & A user's custom command finished executing. Includes the command's code, the message content before it was run, and the updated command. \\
 \hline
 \verb|snippet_resolution| & 102,730 & A user's custom argument shortcut finished executing. Includes the shortcut's code, the argument list before it was run, and the updated argument list. \\
 \hline
 \verb|combat_start| & 24,748 & A user started combat in a channel. \\
 \hline
 \verb|combat_end| & 23,469 & A user ended combat in a channel. (Combats left inactive for over a month are automatically ended without emitting an event.) \\
 \hline
 \verb|button_press| & 21,756 & A user pressed a button associated with a game effect, such as the "Stand Up" button associated with the Prone effect. \\
 \hline
\end{tabularx}
\normalsize
\caption{The number of events in the dataset.}
\label{table:events}
\end{table}

Table \ref{table:events} lists all the events in the dataset along with their descriptions. See \url{https://github.com/avrae/avrae/blob/v4.2.2/cogs5e/initiative/upenn_nlp.py} for each event's schema.

\section{Action Attributes}
\label{sec:action-events}

Actions consist of a tree of \emph{effects}, such as rolling to hit, dealing damage, or rolling a saving throw. Table \ref{table:action-attrs} lists many relevant effects and the attributes available in each. Further documentation about actions is available at \url{https://avrae.readthedocs.io/en/latest/automation_ref.html} and the full definition is available at \url{https://github.com/avrae/avrae/blob/v4.2.2/cogs5e/models/automation/results.py}.

\begin{table}[h]
\centering
\begin{tabularx}{\textwidth}{|l|l|X|} 
 \hline
 {\bf Effect} & {\bf Attribute} & {\bf Description} \\
 \hline\hline
 attack & did\_hit & {\em Whether or not an attack roll hit its target.} \\
 \cline{2-3}
 & did\_crit & {\em Whether or not an attack roll was a critical hit.} \\
 \hline
 save & dc & {\em The difficulty class of this saving throw.} \\
 \cline{2-3}
 & ability & {\em The ability this saving throw uses.} \\
 \cline{2-3}
 & did\_save & {\em Whether the saving roll was successful.} \\
 \hline
 damage & damage & {\em The total amount of damage dealt to a target.} \\
 \cline{2-3}
 & in\_crit & {\em Whether the damage dealt was part of a critical hit.} \\
 \hline
 temphp & amount & {\em The total amount of temporary hit points granted to a target.} \\
 \hline
 ieffect & effect & {\em A temporary game effect was placed on a target (e.g. Prone).} \\
 \hline
 remove\_ieffect & effect & {\em A temporary game effect was removed from a target.} \\
 \hline
 check & skill\_name & {\em The name of a skill the target rolled for a check.} \\
 \cline{2-3}
 & dc & {\em The difficulty class of this ability check, if applicable.} \\
 \cline{2-3}
 & did\_succeed & {\em Whether the check was against a DC and the target succeeded.} \\
 \cline{2-3}
 & contest\_roll & {\em The result of the caster's contesting ability check, if applicable.} \\
 \cline{2-3}
 & contest\_did\_tie & {\em Whether the both actors rolled the same result in a contest.} \\
 \hline
\end{tabularx}
\caption{The attributes associated with each effect in an action.}
\label{table:action-attrs}
\end{table}

\section{Utterance To Command Generation Perturbation Details}
\label{perturbation-sec}
For the perturbation experiments, we selected 2 specific scenarios, one in which the player chooses to cast a bardic inspiration spell to target a single member of their party and the other where the player chooses to cast Fireball, perhaps the most canonical spell in D\&D. For both of these scenarios; we took the combat state and prompt from their respective unit tests and then pertubed them in order to test the ability of the model to react to various modifications in input. We seek to study the model responses specifically to the following scenarios
\begin{itemize}
    \item Targeting and association - can the model pick up intended targets from nicknames/character classes/races in the input. Eg, Can it determine that "Inspires the druid" and "Inspires Noxxis" should result in targeting the same character (if Noxxis is a druid)?
    \item Can it recognize a spell from a creative description of its effects?
    \item Does it attend to spells in the prepared spell list?

While we do not perform exhaustive quantitative analysis, a preliminary analysis indicates that the Finetuned model that includes the state information can react to changes in the prepared spell list. 
\begin{lstlisting}
    "Actors:
        - OR2 (Orc) <9/15 HP; Injured>
        - Reef (Variant Human; Sorcerer 1/Bard 2) <19/25 HP; Injured>
        - KO3 (Kobold) <5/5 HP; Healthy>
        - Rahotur (Mountain Dwarf; Barbarian 6) <77/77 HP; Healthy>
        - Calti Xihooda (Lizardfolk; Druid 6) <45/45 HP; Healthy>
        - Noxxis Blazehammer (Hill Dwarf; Cleric 7) <59/59 HP; Healthy>
        - OR1 (Orc) <13/15 HP; Injured>
        - KO1 (Kobold) <3/5 HP; Injured>
        - KO2 (Kobold) <5/5 HP; Healthy>
        - OR3 (Orc) <15/15 HP; Healthy>
        - OR4 (Orc) <15/15 HP; Healthy>
    Current:
        Name: Noxxis Blazehammer
        Class: Cleric 7
        Race: Hill Dwarf
        Attacks: Warhammer, 2-Handed Warhammer, Unarmed Strike
        Spells: Fireball, Death Ward, Word of Radiance, Hold Person, Spiritual Weapon, Revivify, Augury, Scorching Ray, Light, Healing Word, Spirit Guardians, Guidance, Burning Hands, Faerie Fire, Guiding Bolt, Flaming Sphere, Thaumaturgy, Cure Wounds, Bless, Protection from Evil and Good, Daylight, Wall of Fire, Sacred Flame, Guardian of Faith
        Actions: Channel Divinity, Warding Flare, Channel Divinity: Radiance of the Dawn, War Caster, Channel Divinity: Turn Undead, Destroy Undead, Harness Divine Power
        
        *Noxxis invokes divine anger of his deity, coalescing it into a gout of flame that he launches towards the orcs*",
\end{lstlisting}
This prompt generates the command \emph{!cast fireball -t OR1 -t OR2 -t OR3 -t OR4} but if the Fireball spell is removed from the prompt, it generates the command \emph{!cast "burning hands" -t or1 -t or2 -t or3 -t or4} Similarly, in the case of the Bardic Inspiration example, it's able to replace Bardic Inspiration with Healing Word. The model seems to be able to reliably differentiate between healthy and injured enemies - asking the model to cast Fireball at the injured enemies generates the appropriate command. It also seems to be able to target based on character classes and races. However, it does not always target the correct number of enemies - asking the model to target "2 injured orcs" leads to targeting all the injured orcs. Similarly, asking the model to target based on party roles fails; asking the model to target "the casters" does not correctly target the spellcasters in the party.
While these results are promising, we leave exhaustive quantitative evaluation to future work.
\end{itemize}

\section{Full State to Narration Prompt}
\label{sec:full-sta-nar-prompt}

\begin{lstlisting}
History:
Player 3: (thunder, if it matters?)
Player 3: As the blast hits the hag, another Wild Magic Surge bursts from the gobbo
Player 3: And they turn blue. They look like a Verdan now, and they feel slight amounts of shame from it
Player 3: But, the day goes on, and the reach out with a spectral hand to backhand slap the hag with... Chill Touch, imbuing it with the Tides of Chaos for a little extra sting
Player 4: Well, looks like he got up fine by himself...

There's more bashing to do. Raising the head of the mace, she brings it down once again.
---

Actors:
- Verity Silverdust (Halfling; Rogue 3) <18/18 HP; Healthy> [Mage Armor]
- Nitar (Variant Human; Barbarian 3) <1/35 HP; Critical> [Frightened, Wildhunt Shifting, Rage]
- Bartholomew (Goblin; Sorcerer 3) <23/23 HP; Healthy> [Wild Resistance, Chilling Touch]
- Alexsandra (Astral Elf; Cleric 3) <15/15 HP; Healthy>
- Keya (Custom Lineage; Fighter 2/Warlock 1) <24/24 HP; Healthy> [Hexblade's Curse, Hex, Hexing]
- Mozzie Urahaka (Dhampir; Artificer 1/Wizard 2) <22/22 HP; Healthy> [Mind Splinter]
- SH1 (Sea Hag) <2/52 HP; Critical> [Hexblade's Cursed, Chill Touch, Hexed]

Targets:
- SH1 (Sea Hag) <2/52 HP; Critical> [Hexblade's Cursed, Chill Touch, Hexed]

Description:
A timeless woman with flowing star-speckled hair that fades between nebula violet and empty black, she wears and carries exotic armor and weaponry: most of all being the amulet of a horned clawed cyclopic serpent swallowing its own tail. She manages the altar and gravestones at the cemetery and ensures that no desecration comes to those under her care. When spoken to, the gravekeeper has an odd aura about her, being generally much too wide-eyed and profound for simple small-talk.
---
Name: Alexsandra
Class: Cleric 3
Race: Astral Elf
Attacks: Crossbow, light, Mace, Unarmed Strike
Actions: Channel Divinity: Radiance of the Dawn, Channel Divinity: Turn Undead, Harness Divine Power, Starlight Step, Warding Flare
Spells: Dancing Lights, Guidance, Light, Sacred Flame, Spare the Dying, Thaumaturgy, Bless, Burning Hands, Command, Faerie Fire, Healing Word, Sanctuary, Blindness/Deafness, Flaming Sphere, Gentle Repose, Lesser Restoration, Scorching Ray, Silence

Aleksandra attacks with a Mace!
Aleksandra attacked SH1 but missed.
\end{lstlisting}

\section{Full Utterance to Command Prompt}
\label{sec:full-utt-cmd-prompt}

\begin{lstlisting}
Actors:
- Filgo Bitterfoot (Mountain Dwarf; Fighter 5) <43/43 HP; Healthy>
- DW1 (Dire Wolf) <25/37 HP; Injured>

Current:
Name: Filgo Bitterfoot
Class: Fighter 5
Race: Mountain Dwarf
Attacks: Greataxe, Longsword, Longbow, Handaxe
Actions: Second Wind, Action Surge

Filgo swings his axe at the wolf! "Raaaargh!"
\end{lstlisting}

For this example, the true command is:
\begin{lstlisting}
!a greataxe -t dw1
\end{lstlisting}

\section{Human Evaluation Interface}
\label{sec:eval-interface}

\subsection{Signup Form}

To assess each rater's familiarity with Avrae and D\&D, we asked each the following questions:

\begin{enumerate}
    \item Have you ever played the fifth edition of Dungeons and Dragons (D\&D 5e) using the Avrae Discord Bot before?
    \item If so, roughly how long have you been playing using Avrae?
    \item Have you ever been the Dungeon Master of a D\&D 5e game using Avrae?
    \item If so, roughly how long have you been DMing using Avrae?
    \item Have you used any of these commands on the Avrae Discord Bot?
    \begin{enumerate}
        \item !import, !check, !save, !action, !cast
        \item !init begin, !init next, !init join, !init action
    \end{enumerate}
    \item Have you ever played on a "play-by-post" D\&D Discord?
    \item What's one feature you always wished was in the Avrae Discord Bot?
\end{enumerate}

\subsection{Evaluation Mockup}
In this task, you will see part of a play-by-post D\&D combat using Avrae in the form of Discord messages leading up to an Avrae action. The caster's description and current initiative list are listed along with the Discord messages. The messages that are shown as context are real messages from players. Your job is to read the context and then rate different responses for the dungeon master's narration of the action. Please note that the context you are given represents only a part of the players' past conversations/interactions with one another during the game. 

\subsubsection*{Caster's Description}
\begin{quote}
Description: 5'10 (180cm) | 180 lb. | Chromatic Dragonborn | Fighter (Battle Master)/Bard

Young, lean but strong overall build. They're a blue chromatic dragonborn who's always seen in armor and formal decorated robes. With silvery blond hair that is usually hidden behind a helm. Umbrage has many choices of weaponry, not one to pick or choose when it comes to the field of battle. But his most favored would be that horn of his. A rustic and old warhorn, The ivory it's made from is something unusual, even going so far as to be able to tap into the wave around him by chance.

With the passing battles, many new scars are shown upon scales. But the only one that bothers him the most and that is always is kept hidden. Is the injury found upon his neck. The cause must have been something heavy enough to leave a lasting imprint, but Umbrage would never tell what it was. Shocking anyone who gets too close to that reverse scale.
\end{quote}

\subsubsection*{Initiative List}
\begin{lstlisting}
- DD6 (Death Dog) <11/39 HP; Bloodied> [baki]
- Katherine (Dhampir; Rogue 2/Blood Hunter 3) <46/46 HP; Healthy>
- Holawynn Meitorin (Satyr; Cleric 5) <48/48 HP; Healthy>
- DD1 (Death Dog) <39/39 HP; Healthy> [Umbrage]
- DD2 (Death Dog) <-19/39 HP; Dead> [yala]
- Kaska (Leonin; Ranger 7/Paladin 3) <114/114 HP; Healthy>
- Baki (Beast of the Land) <33/40 HP; Injured> [Maul, Primal Bond, Poisoned]
- DD7 (Death Dog) <39/39 HP; Healthy> [Lytrha]
- Umbrage (Chromatic Dragonborn; Fighter 3/Bard 4) <56/63 HP; Injured> [Cause Fear]
- Lythra (Half-Orc; Ranger 3/Rogue 2) <42/42 HP; Healthy>
- Yala the Wanderer (Goblin; Bard 5/Rogue 1) <39/39 HP; Healthy>
- DD8 (Death Dog) <39/39 HP; Healthy> [40 feet, Frightened (Cause Fear)]
- DD3 (Death Dog) <39/39 HP; Healthy> [40 feet, Frightened (Cause Fear)]
- DD5 (Death Dog) <39/39 HP; Healthy> [40 feet, Frightened (Cause Fear)]
- DD4 (Death Dog) <39/39 HP; Healthy> [40 feet]
\end{lstlisting}

\subsubsection*{Context}
\begin{lstlisting}
Avrae | 3 minutes ago
Initiative 12 (round 1): Umbrage ()

Umbrage <56/63 HP> (AC 20)
> Resistances: lightning
> Immunities: critical hits, diseas

Player 0
 a minute ago
"Dogs... so many of them making it more annoying then anything." Umbrage didn't want to waste anything big. So he pulled out his pipes, taking the attack of opportunity as he would start to play.

"Fear me!"

Player 0 | now
!i cast "Cause Fear" -dc 15 -t dd1 -t dd3 -t dd4 -t dd5 -t dd6 -t dd7 -t dd8 sadv -i -title "Pipes of Haunting!" -thumb https://cdn.discordapp.com/attachments/979075315706003459/1005867411385364480/giphy_6.gif

Avrae | now
Pipes of Haunting!
Meta
DC: 15

DD1
WIS Save: 2d20kh1 (15, 12) + 1 = 16; Success!

DD3
WIS Save: 2d20kh1 (5, 2) + 1 = 6; Failure!
Effect: Frightened (Cause Fear) [10 rounds] (Resist Fear; Parent: Cause Fear)

DD4
WIS Save: 2d20kh1 (9, 16) + 1 = 17; Success!

DD5
WIS Save: 2d20kh1 (2, 8) + 1 = 9; Failure!
Effect: Frightened (Cause Fear) [10 rounds] (Resist Fear; Parent: Cause Fear)

DD6
WIS Save: 2d20kh1 (19, 18) + 1 = 20; Success!

DD7
WIS Save: 2d20kh1 (20, 17) + 1 = 21; Success!

DD8
WIS Save: 2d20kh1 (9, 3) + 1 = 10; Failure!
Effect: Frightened (Cause Fear) [10 rounds] (Resist Fear; Parent: Cause Fear)

Effect
You awaken the sense of mortality in one creature you can see within range. A construct or an undead is immune to this effect. The target must succeed on a Wisdom saving throw or become frightened of you until the spell ends. The frightened target can repeat the saving throw at the end of each of its turns, ending the effect on itself on a success.
\end{lstlisting}

\subsubsection*{Does the response make sense?}

Use your common sense here. The response "makes sense" if:
\begin{enumerate}
    \item it is cohesive as a standalone statement, 
    \item consistent with the rules of the game, 
    \item and it is a plausible narration given the prior context (initiative list and last actions taken).
\end{enumerate}
If anything seems off—not fluent, confusing, illogical, out of context, or wrong according to the rules of D\&D —then choose No. If in doubt about a response, choose No.

\begin{table}[H]
\centering
\begin{tabularx}{\textwidth}{X c c} 
  & Yes & No \\
  The music was certainly not pleasant, but it seemed to have an effect on the dogs. They became panicked for a moment, but were able to maintain their aggression. It was just enough for the party to act. & $\square$ & $\square$ \\
  Cuco proceeds to try and scare the ducks. So far it wasn't working. & $\square$ & $\square$ \\
  Moving back into the melee with the one he was originally fighting. As he drew ready. & $\square$ & $\square$ \\
  As the first dog fell, the others seemed to flee once more in pure terror of the seemingly unending tunes of the pipes. & $\square$ & $\square$ \\
  The dogs were in the middle of a panic. But still they fought on! & $\square$ & $\square$ \\
\end{tabularx}
\end{table}

\subsubsection*{Is the response specific?}
In other words, do you think that the response accurately narrates the last action the character actually took and its results?

The response is "specific" if it flows logically from the specific action and result taken by the character, in the greater context provided.

Note: It is possible for a response to "make sense" (due to being cohesive, consistent and plausible in and of itself), but be marked "not specific" when it is not a logical next step in the overall game progression.

Note: "Specific" for the purposes of this task does not have to do with how detailed the response is per se; a response can be fairly general in its language, but still qualify as "specific" when it is a logical next step in the overall game progression.

\begin{table}[H]
\centering
\begin{tabularx}{\textwidth}{X c c} 
  & Yes & No \\
  As the first dog fell, the others seemed to flee once more in pure terror of the seemingly unending tunes of the pipes. & $\square$ & $\square$ \\
  Cuco proceeds to try and scare the ducks. So far it wasn't working. & $\square$ & $\square$ \\
  Moving back into the melee with the one he was originally fighting. As he drew ready. & $\square$ & $\square$ \\
  The dogs were in the middle of a panic. But still they fought on! & $\square$ & $\square$ \\
  The music was certainly not pleasant, but it seemed to have an effect on the dogs. They became panicked for a moment, but were able to maintain their aggression. It was just enough for the party to act. & $\square$ & $\square$ \\
\end{tabularx}
\end{table}

\subsubsection*{How interesting is the response? (10 is best)  }
Rank a response as more “Interesting” if the response would likely catch someone's attention or arouse curiosity in the game; or it is insightful, creative, or witty with respect to the game. If the response is monotonous and predictable, then rank it lower. If anything seems off—not fluent, confusing, illogical, out of context, or wrong according to the rules of D\&D —then rank it lower.

\begin{table}[H]
\centering
\newcolumntype{L}{>{\raggedright\arraybackslash}X}
\newcolumntype{R}{>{\raggedleft\arraybackslash}X}
\newcolumntype{Y}{>{\centering\arraybackslash}X}
\newcolumntype{Z}{>{\hsize=\dimexpr10\hsize+18\tabcolsep+9\arrayrulewidth\relax}X}  
\begin{tabularx}{\textwidth}{ LYYYYYYYYR }
  \multicolumn{5}{ l }{Less Interesting} & \multicolumn{5}{ r }{More Interesting} \\
  1 & 2 & 3 & 4 & 5 & 6 & 7 & 8 & 9 & 10 \\
  \hline
  \multicolumn{10}{Z}{The music was certainly not pleasant, but it seemed to have an effect on the dogs. They became panicked for a moment, but were able to maintain their aggression. It was just enough for the party to act.} \\
  $\bigcirc$ & $\bigcirc$ & $\bigcirc$ & $\bigcirc$ & $\bigcirc$ & $\bigcirc$ & $\bigcirc$ & $\bigcirc$ & $\bigcirc$ & $\bigcirc$ \\
  \hline
  \multicolumn{10}{Z}{The dogs were in the middle of a panic. But still they fought on!} \\
  $\bigcirc$ & $\bigcirc$ & $\bigcirc$ & $\bigcirc$ & $\bigcirc$ & $\bigcirc$ & $\bigcirc$ & $\bigcirc$ & $\bigcirc$ & $\bigcirc$ \\
  \hline
  \multicolumn{10}{Z}{Cuco proceeds to try and scare the ducks. So far it wasn't working.} \\
  $\bigcirc$ & $\bigcirc$ & $\bigcirc$ & $\bigcirc$ & $\bigcirc$ & $\bigcirc$ & $\bigcirc$ & $\bigcirc$ & $\bigcirc$ & $\bigcirc$ \\
  \hline
  \multicolumn{10}{Z}{As the first dog fell, the others seemed to flee once more in pure terror of the seemingly unending tunes of the pipes.} \\
  $\bigcirc$ & $\bigcirc$ & $\bigcirc$ & $\bigcirc$ & $\bigcirc$ & $\bigcirc$ & $\bigcirc$ & $\bigcirc$ & $\bigcirc$ & $\bigcirc$ \\
  \hline
  \multicolumn{10}{Z}{Moving back into the melee with the one he was originally fighting. As he drew ready.} \\
  $\bigcirc$ & $\bigcirc$ & $\bigcirc$ & $\bigcirc$ & $\bigcirc$ & $\bigcirc$ & $\bigcirc$ & $\bigcirc$ & $\bigcirc$ & $\bigcirc$ \\
\end{tabularx}
\end{table}

\section{Human Evaluation Significance}
\label{sec:human-eval-sig}

We use the Student's t-test to calculate significance for our three rated dimensions. The results are tabulated below, with \textbf{bold} indicating $p < 0.001$, \emph{italics} indicating $p < 0.01$, and \textdagger{} indicating $p < 0.05$:

\begin{table}[H]
\centering
Significance of Sense (p-value)
\begin{tabular}{l|ccccc}
  & FIREBALL-FULL & FIREBALL-SHORT & COMMAND & DIALOG \\
  \hline
  FIREBALL-SHORT & 0.2896 & - & - & - \\
  COMMAND        & \textbf{0.0002} & \emph{0.0046} & - & - \\
  DIALOG         & \textbf{0.0000} & \textbf{0.0000} & 0.1410 & - \\
  Human          & 0.5995 & 0.6179 & \emph{0.0014} & \textbf{0.0000} \\
\end{tabular}
\end{table}

\begin{table}[H]
\centering
Significance of Specific (p-value)
\begin{tabular}{l|ccccc}
  & FIREBALL-FULL & FIREBALL-SHORT & COMMAND & DIALOG \\
  \hline
  FIREBALL-SHORT & 0.7412 & - & - & - \\
  COMMAND        & \emph{0.0079} & \emph{0.0025} & - & - \\
  DIALOG         & \textbf{0.0000} & \textbf{0.0000} & \emph{0.0060} & - \\
  Human          & 0.8670 & 0.8651 & \emph{0.0059} & \textbf{0.0000} \\
\end{tabular}
\end{table}

\begin{table}[H]
\centering
Significance of Interesting (p-value)
\begin{tabular}{l|ccccc}
  & FIREBALL-FULL & FIREBALL-SHORT & COMMAND & DIALOG \\
  \hline
  FIREBALL-SHORT & 0.0602 & - & - & - \\
  COMMAND        & 0.5189 & 0.1996 & - & - \\
  DIALOG         & 0.0710 & \textbf{0.0001} & 0.0227\dagsuper & - \\
  Human          & 0.1288 & 0.7277 & 0.4153 & \emph{0.0019} \\
\end{tabular}
\end{table}

\section{Example of Full State to Narration Context}
\label{sec:sta-nar-example}

The following is the full prompt provided to the FIREBALL-FULL model in the example provided in Table \ref{table:examples-sta-nar}.

\begin{lstlisting}
History:
Player 1: *Holawynn would back up 35 if she could. The satyr knew where she should be standing during this fight. She starts up with a twilight flame to conserve slots.*
Player 1: *It misses out of sheer unluck. A shame that it was also kinda crap.*
Player 0: ```The hounds charged. Each managing a singular bite on their targets. It would seem they were all wanting to eat some tender flesh of the adventures who passed through their masters lair!```
Player 2: *Kaska looked about the combat and swung her weapon to Yala's aid, Baki attacking the dog that wanted to eat his bacon.*
Player 0: "Dogs... so many of them making it more annoying then anything." *Umbrage didn't want to waste anything big. So he pulled out his pipes, taking the attack of opportunity as he would start to play.* 
"**Fear me!**"
---

Actors:
- DD6 (Death Dog) <11/39 HP; Bloodied> [baki]
- Katherine (Dhampir; Rogue 2/Blood Hunter 3) <46/46 HP; Healthy>
- Holawynn Meitorin (Satyr; Cleric 5) <48/48 HP; Healthy>
- DD1 (Death Dog) <39/39 HP; Healthy> [Umbrage]
- DD2 (Death Dog) <-19/39 HP; Dead> [yala]
- Kaska (Leonin; Ranger 7/Paladin 3) <114/114 HP; Healthy>
- Baki (Beast of the Land) <33/40 HP; Injured> [Maul, Primal Bond, Poisoned]
- DD7 (Death Dog) <39/39 HP; Healthy> [Lytrha]
- Umbrage (Chromatic Dragonborn; Fighter 3/Bard 4) <56/63 HP; Injured> [Cause Fear]
- Lythra (Half-Orc; Ranger 3/Rogue 2) <42/42 HP; Healthy>
- Yala the Wanderer (Goblin; Bard 5/Rogue 1) <39/39 HP; Healthy>
- DD8 (Death Dog) <39/39 HP; Healthy> [40 feet, Frightened (Cause Fear)]
- DD3 (Death Dog) <39/39 HP; Healthy> [40 feet, Frightened (Cause Fear)]
- DD5 (Death Dog) <39/39 HP; Healthy> [40 feet, Frightened (Cause Fear)]
- DD4 (Death Dog) <39/39 HP; Healthy> [40 feet]

Targets:
- DD1 (Death Dog) <39/39 HP; Healthy> [Umbrage]
- DD3 (Death Dog) <39/39 HP; Healthy> [40 feet, Frightened (Cause Fear)]
- DD4 (Death Dog) <39/39 HP; Healthy> [40 feet]
- DD5 (Death Dog) <39/39 HP; Healthy> [40 feet, Frightened (Cause Fear)]
- DD6 (Death Dog) <11/39 HP; Bloodied> [baki]
- DD7 (Death Dog) <39/39 HP; Healthy> [Lytrha]
- DD8 (Death Dog) <39/39 HP; Healthy> [40 feet, Frightened (Cause Fear)]

Description: __**5'10 (180cm) | 180 lb. | Chromatic Dragonborn | Fighter (Battle Master)/Bard**__
> Young, lean but strong overall build. They're a blue chromatic dragonborn who's always seen in armor and formal decorated robes. With silvery blond hair that is usually hidden behind a helm. Umbrage has many choices of weaponry, not one to pick or choose when it comes to the field of battle. But his most favored would be that horn of his. A rustic and old warhorn, The ivory it's made from is something unusual, even going so far as to be able to tap into the wave around him by chance.
>
> With the passing battles, many new scars are shown upon scales. But the only one that bothers him the most and that is always is kept hidden. Is the injury found upon his neck. The cause must have been something heavy enough to leave a lasting imprint, but Umbrage would never tell what it was. Shocking anyone who gets too close to that reverse scale.
**Personality Traits**
I'm haunted by memories of war. I can't get the images of violence out of my mind.
I'm full of inspiring and cautionary tales from my military experience relevant to almost every combat situation.
I can stare down a hell hound without flinching.
**Ideals**
Might. In life as in war, the stronger force wins. (Evil)
**Bonds**
My honor is my life.
I'll never forget the crushing defeat my company suffered or the enemies who dealt it.
**Flaws**
I made a terrible mistake in battle that cost many lives-and I would do anything to keep that mistake secret.
**Alignment**
NE
---
Name: Umbrage
Class: Fighter 3/Bard 4
Race: Chromatic Dragonborn
Attacks: Javelin of Lightning, Light Hammer, +1, Longsword, +1, 2-Handed Longsword, +1, Warhammer, +1, Unarmed Strike, Keoghtom's Ointment
Spells: Ice Storm, Hold Person, Sleet Storm, Vicious Mockery, Faerie Fire, Ray of Frost, Thunderclap, Healing Word, Bane, Dissonant Whispers, Silence, Sleep, Prestidigitation
Actions: Maneuvers: Commander's Strike, Superiority Dice, Action Surge, Maneuvers: Bait and Switch (Self), Maneuvers: Bait and Switch (Target), Combat Inspiration, Song of Rest, Martial Adept, Maneuvers: Distracting Strike, Second Wind, Bardic Inspiration, Lightning Breath Weapon, Maneuvers: Ambush
Effects: Cause Fear

Pipes of Haunting!
DD1 rolled a Wisdom save and succeeded.
DD3 rolled a Wisdom save but failed.
DD3 gained Frightened (Cause Fear).
DD4 rolled a Wisdom save and succeeded.
DD5 rolled a Wisdom save but failed.
DD5 gained Frightened (Cause Fear).
DD6 rolled a Wisdom save and succeeded.
DD7 rolled a Wisdom save and succeeded.
DD8 rolled a Wisdom save but failed.
DD8 gained Frightened (Cause Fear).
\end{lstlisting}

\end{document}